\documentclass{llncs}
\usepackage{makeidx}  
\usepackage{times}
\usepackage[english]{babel}
\usepackage{amsmath}
\usepackage{graphicx}
\usepackage{caption}

\begin{document}
\frontmatter          
\pagestyle{headings}  

\title{R-Clustering for Egocentric Video Segmentation}

\titlerunning{Egocentric Video Segmentation}  
%
\author{Estefania Talavera \inst{1,2} \and Mariella Dimiccoli \inst{1,3} \and Marc Bola{\~n}os \inst{1} \and \\ Maedeh Aghaei \inst{1} \and Petia Radeva \inst{1,3}}
\authorrunning{Estefania Talavera et al.} 
%
\tocauthor{Estefania Talavera, Mariella Dimiccoli, Marc Bola{\~n}nos, Maedeh Aghaei, Petia Radeva}
\institute{Universitat de Barcelona, Barcelona, Spain\\
\and
University of Groningen, Groningen, Netherlands\\
\and
Computer Vision Center, Barcelona, Bellaterra, Spain\\
\email{\small etalavera@ub.edu, mariella.dimiccoli@cvc.uab.es, marc.bolanos@ub.edu, maghaeigavari@ub.edu, petia.ivanova@ub.edu}
}

 \maketitle              

\vspace{-2em}

\begin{abstract}

In this paper, we present a new method for egocentric video temporal segmentation based on integrating a statistical mean change detector and agglomerative clustering(AC) within an energy-minimization framework. Given the tendency of most AC methods to oversegment video sequences when clustering their frames, we combine the clustering with a concept drift detection technique (ADWIN) that has rigorous guarantee of performances. ADWIN serves as a statistical upper bound for the clustering-based video segmentation. We integrate both techniques in an energy-minimization framework that serves to disambiguate the decision of both techniques and to complete the segmentation taking into account the temporal continuity of video frames descriptors. We present experiments over egocentric sets of more than 13.000 images acquired with different wearable cameras, showing that our method outperforms state-of-the-art clustering methods.

\end{abstract}

\section{Introduction}
\vspace*{-0.25cm}Lifestyle behaviour is closely related with health outcomes, in particular, to noncommunicable diseases such as obesity and depression, that represent a major burden in developed countries.
\begin{figure}[hb!]
\includegraphics[width=\linewidth,height=10em]{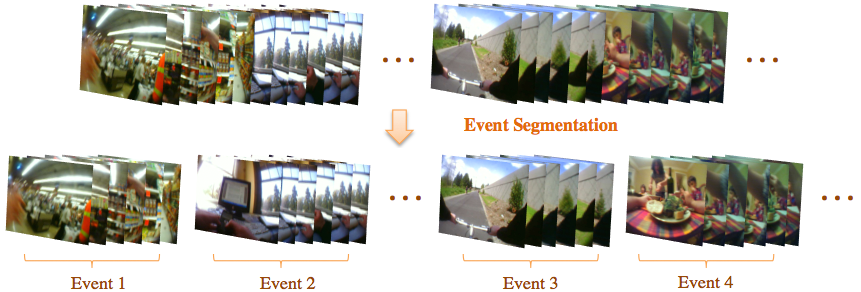}
\caption{Example of temporal segmentation of a SenseCam sequence.}
\label{fig:segmentation}
\vspace{-2em}
\end{figure}
A promising way towards studying one's lifestyle is through the use of wearable cameras, able to digitally capture a person's everyday activities into the so called lifelogging. However, the automatic recognition of daily routines using wearable devices is very challenging due to the huge amount of collected data (up to 3.000 images per day). Moreover, daily routines are typically composed of many complex events, with a large variability depending on factors such as time, location and individual. This work proposes an algorithm for grouping similar temporally adjacent images into segments, providing a structure to egocentric videos, that is important for further analysis such as video summarization and analysis. Considering the environment as a strong characteristic of an event, these segments are supposed to characterize different environments in which the camera wearer acts (Fig. \ref{fig:segmentation}).



Previous methods for egocentric temporal segmentation can be classified into two broad classes, depending on whether they rely on low-level or high-level features. Basically, the former class has as focus what the {\em wearer sees} and uses as image representation features that are able to capture the characteristics of the environment around the wearer, such as color and texture; the latter class focuses on what the {\em wearer does} and thus uses as image representation high-level concepts such as objects and activities.

Early works based on low-level features include the one of Doherty et al. \cite{Doherty:2008}, which is based on the use of MPEG-7 descriptors for image representation that are available from the sensor, and the one of \cite{kmean}, that uses a time-constrained k-means algorithm based on color descriptors.
Recent methods focus on motion-based features. Usually, optical flow is used to distinguish between static, moving the head/camera and {\em in-transit} frames \cite{egovideo,bolanos2014video}. This classification offers a segmentation that focuses in the activities and movements performed by the user, but is prone to fail when the environment changes while performing the same activity (e.g. the user is in transit but, first gets out of their workplace, then is walking on the street and finally enters to the underground). To focus on long-term activities, Poleg et al. \cite{polegtemporal} proposed the use of integral motion, which is closely related to wearers' activity. By integrating the instantaneous displacements at fixed image patches, the variations due to head rotation are eliminated, since their mean is practically zero, leaving only the consistent displacement caused by forward motion. Methods based on motion analysis assume high temporal resolution, but the temporal resolution of many lifelogging devices, such those considered in this paper, is very low.

Works based on high-level features are generally more recent. In \cite{track}, first  important people and objects are discovered by measuring their interaction with the camera wearer and then the frames which reflect the key objects happening are selected. In \cite{graphmod}, the authors propose a summarization tool based on  analysis of video structures and video highlights. By emphasizing on both the content balance and the perceptual quality of the summary, the authors employ a normalized cut algorithm to globally and optimally partition a video into clusters. Furthermore, in \cite{spatemp}, the authors present a video segmentation approach based on the study of spatio-temporal activities within the video, that leads to a visual activity estimation by measuring the number of interest points, jointly obtained in the spatial and temporal domains.  

In this paper, we rely on low-level features. Our approach is a {\em Graph-Cut (GC) extension} technique  \cite{boykov2001fast,bolanos2014video} that takes advantage of two methods having complementary properties: ADWIN \cite{bifet2007learning} - a concept drift technique for mean change detection that is highly precise, but usually leads to temporal under-segmentation; and  agglomerative clustering (AC), which usually has a high recall, but leads to temporal  over-segmentation. Our approach, that we call {\bf R-Clustering}, {\em regularizes} the over-segmentation of the AC through the upper bound provided by ADWIN.  Based on the excellent accuracy achieved recently for classification in a variety of computer vision tasks \cite{NIPS2012_4824,Goodfellow2014Multi}, we use Convolutional Neural Network (CNN) vector activation over the entire image as a global image feature descriptor. CNN features are able to focus just in the environment appearance and do not need to rely on a motion information that, would be unfeasible to estimate reliably  taking into account the very low temporal resolution of the wearable devices we considered (up to 3fpm). As an example of application, we illustrate the utility of the proposed method for the detection of social events. 
In the next section, we detail the proposed approach. In section \ref{sec:results}, we discuss experimental results and, finally,  in section \ref{sec:conclusions} we draw some conclusions.

\section{The R-Clustering Approach for Temporal Video Segmentation}
\label{sec:approach}
\vspace*{-0.25cm}Due to the low-temporal resolution of egocentric videos, as well as to the camera wearer's motion, temporally adjacent egocentric images may be very dissimilar between them. Hence, we need robust techniques to group them and extract meaningful video segments. In the following, we detail each step of our approach that relies on an AC regularized by a robust change detector within  a GC framework.  

 

\vspace*{-0.5cm}
\subsubsection{Clustering methods:}
The AC method follows a general bottom-up clustering procedure, where the criterion for choosing the pair of clusters to be merged in each step is based on the distances among the image features. The inconsistency between clusters is defined through the {\em cut} parameter. In each iteration, the most similar pair of clusters are merged and the similarity matrix is updated until no more consistent clustering are possible. We chose the Cosine Similarity to measure the distance between frames features, since it is a widely used measure of cohesion within clusters, specially in high-dimensional positive spaces \cite{Tan:2005:IDM:1095618}. However, due to the lack of incidence for determining the clustering parameters, the final result is usually over-segmented.

\vspace*{-0.5cm}
\subsubsection{Statistical bound for the clustering:}
To bound the over-segmentation produced by AC, we propose to model the video as a multivariate data stream and detect changes in the mean distribution through an online learning method called  Adaptative Windowing (\textbf{ADWIN}) \cite{bifet2007learning}. ADWIN works by analyzing the content of a sliding window, whose size is adaptively recomputed according to its rate of change: when the data is stationary the window increases, whereas when the data is statistically changing, the window shrinks. According to ADWIN, whenever two large enough temporally adjacent (sub)windows of the data, say $W_1$ and $W_2$, exhibit distinct enough means, the algorithm concludes that the expected values within those windows are different, and the older (sub)window is dropped.   {\em Large enough} and {\em distinct enough} are defined by the Hoeffding's inequality \cite{Hoeffding1963},  testing if the difference between the averages on $W_1$ and $W_2$ is larger than a threshold, which only depends on a pre-determined confidence parameter $\delta$. The Hoeffding's inequality guarantees rigorously the performance of the algorithm in terms of false positive rate.

\begin{figure}
        \centering
        \hspace*{-0.2em}
        \begin{minipage}[b]{0.33\textwidth}
                \includegraphics[width=\textwidth]{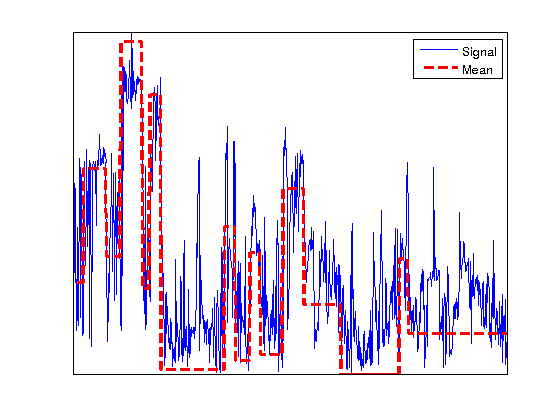}
        \end{minipage}%
        \hspace*{-0.2em}
             \begin{minipage}[b]{0.33\textwidth}
                \includegraphics[width=\textwidth]{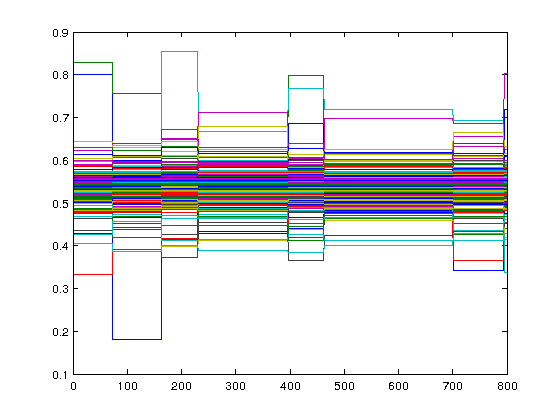}
        \end{minipage}
        \hspace*{-0.2em}
        \begin{minipage}[b]{0.33\textwidth}
                \includegraphics[width=\textwidth]{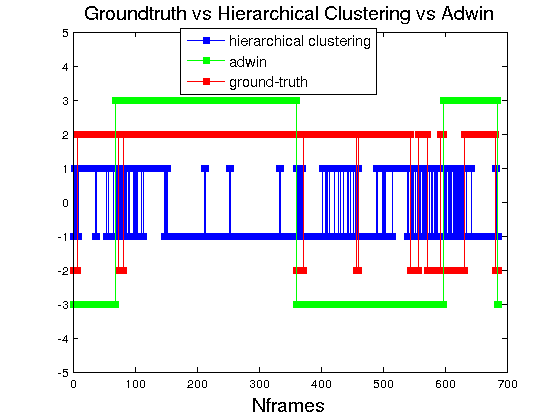} 
        \end{minipage}  
        \vspace*{-2em}
        \caption{Left: change detection by ADWIN on a $1-D$  data stream, where the red line represents the estimated mean of the signal by ADWIN; Center: change detection by ADWIN on a 500-D data stream, where, in each stationary interva,  the mean is depicted with a different color in each dimension; Right: results of the temporal segmentation by ADWIN  (green) vs AC over-segmentation (blue) vs ground-truth shots (red) along the temporal axis (the abscissa). 
        }
        \label{fig:clustering2}
\end{figure}

This method has been recently generalized in \cite{Drozdzal2014} to handle $k-$dimensional data streams by using the mean of the norms. In this case, the bound has been shown to be:
$$\epsilon_{cut} = k^{1/p}\sqrt{\frac{1}{2m}\ln\frac{4}{k\delta'}}$$ where $p$
indicates the $p-$norm, $|W| =|W0|+|W1|$ is the length of $W=W_1\cup W_2$, $\delta'=\frac{\delta}{|W|}$,  and  $m$ is the harmonic mean of $|W0|$ and $|W1|$.
Given a confidence value $\delta$, the higher the dimension $k$ is, the more samples $|W|$ the bound needs to reach assuming the same value of $\epsilon_{cut}$. The higher the norm is used,  the less important is the dimensionality $k$. Since we model the video as a high dimensional multivariate data stream,  ADWIN is unable to predict changes involving a small number of samples, which often characterizes life-logging data, leading to under-segmentation. Moreover, since it considers only the mean change, it is enable to detect changes due to other statistics such as the variance. The ADWIN under-segmentation represents a statistical bound for the AC (see Fig.\ref{fig:clustering2} (right)). We use GC as a framework to integrate both approaches and to regularize the over-segmentation of AC by the statistical bound provided by ADWIN. 
\vspace*{-1.5em}
\vspace*{-1cm}
\subsubsection{Graph-Cut regularization of egocentric videos:} 
GC is an energy-minimization technique that minimizes the energy resulting from a weighted sum of two terms: the {\em unary energy} $U(\_)$, that describes the relationship of the variables to a possible class and the {\em binary energy} $V(\_,\_)$, that describes the relationship between two neighbouring samples (temporally close video frames) according to their feature similarity. GC has the goal to smooth boundaries between similar frames, while attempts to keep the cluster membership of each video frame according to its likelihood. We define the unary energy as a sum of 2 parts ($U_{ac}(f_i)$ and $U_{adw}(f_i)$) according to the likelihood of a frame to belong to segments coming from the AC  and ADWIN. The GC energy to minimize is as follows:
\vspace{-0.5em}
    $$E(f) = \sum_{i} ((1-\omega_1) U_{ac}(f_i) + \omega_1 U_{adw}(f_i) ) + \omega_2 \sum_{i,n \in N_i} \frac{1}{N_i} V_{i,n}(f_i,f_n)$$
where $f_i,i=\lbrace1,...,m \rbrace$ are the set of image features, $N_i$ are the temporal frame neighbours of image $i$, $\omega_1$ and $\omega_2$ ($\omega_1,  \omega2  \in [0,1]$) are the unary and the binary weighting terms respectively. Defining how much weight do we give to the likelihood of each unary term (AC and Adwin, always combining the events split of both methods), and balancing the trade-off between the unary and the pairwise energies, respectively. The minimization is achieved through the max-cut algorithm, leading to a temporal video segmentation with similar frames having as large likelihood as possible to belong to the same event, while maintaining video segment boundaries in neighbouring frames with high feature dissimilarity. 
\vspace*{-0.5cm}
\subsubsection{Features:}
As image representation for both segmentation techniques, we used the CNN features \cite{Jia13caffe}. The CNN features trained on ImageNet \cite{NIPS2012_4824} have demonstrated to be successfully transferred to other visual recognition tasks such as scene classification and retrieval. In this work, we extracted the 4096-D CNN vectors  by using the Caffe \cite{Jia13caffe} implementation trained on ImageNet. Since each CNN feature has a large variation distribution in its value, and this could be problematic when computing distances between vectors, we used a signed root normalization to produce more uniformly distributed data \cite{ZhengWHT14}. First, we apply the function $f(x) = sign(x)|x|^\alpha$ on each dimension and then we $l_2-$normalize the feature vector. In all the experiments, we take $\alpha = 0.5$. Following we apply a PCA dimensionality reduction keeping 95\% of the data variance. Only in the GC pair-wise term we use a different feature pre-processing, where we simply apply a 0-1 data normalization.

\section{Results and validation}
\label{sec:results}
\vspace*{-0.25cm}In this section, we discuss the datasets, the statistical validation measurements, tests and comparison to other methods as well as a possible application of the R-Clustering. 
\vspace*{-0.35cm}
\subsubsection{Data:} 
To evaluate the performance of our method, we used 2 datasets (one public \cite{NIPS} and one made by us), composed of 10 days with a total of 13324 images, acquired by two different wearable devices: SenseCam \cite{sensecam} and Narrative {(http://getnarrative.com/)}. The main differences between the two kind of devices are the frame rate (3 fpm vs 2 fpm) and the lens (fisheye vs normal). The data adquired by the SenseCam contain a larger number of frames per day with a larger field of view and significant deformation and blurring. Both datasets include 5 days each, containing a mix of indoor and outdoor scenes with numerous foreground and background objects. All data has been manually annotated to provide ground-truth segmentation.
\vspace*{-0.35cm}
\subsubsection{Statistical measurements:} As evaluation criterion (following \cite{li2013daily}), we used the F-Measure (FM): $FM= 2(RP)/(R+P)$, where $P$ is the precision $(P=TP/(TP+FP)$, $R$ is the recall $(R=TP/(TP+FN)$ and $TP$, $FP$ and $FN$ are the number of true positives, false positives and false negatives. 
\vspace*{-1em}
\begin{figure}
    \vspace{-1em}
        \centering
        \begin{minipage}[b]{0.33\textwidth}
                \includegraphics[width=\textwidth]{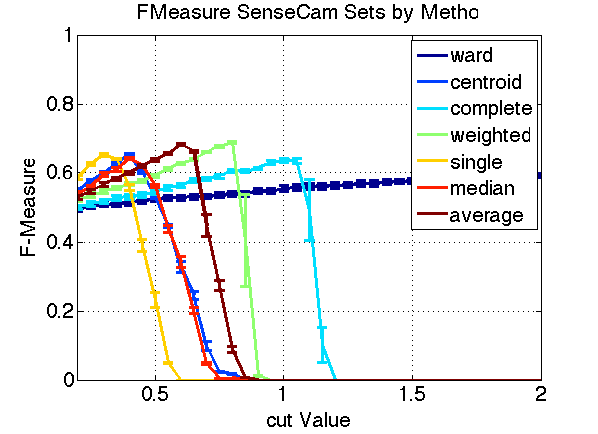}
        \end{minipage}%
        \hspace*{-0.5em}
        \begin{minipage}[b]{0.33\textwidth}
                \includegraphics[width=\textwidth]{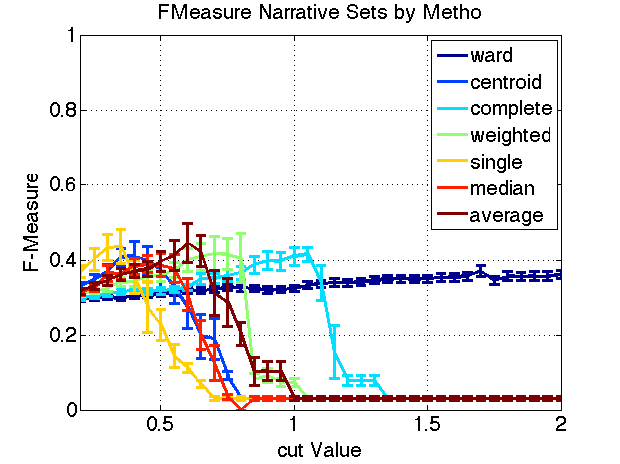}
        \end{minipage}
        \hspace*{-0.5em}
        \begin{minipage}[b]{0.33\textwidth}
                \includegraphics[width=\textwidth]{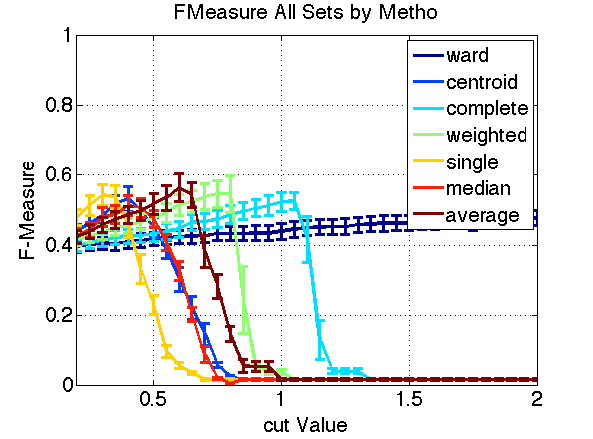}
        \end{minipage}
        \vspace*{-1em}
        \caption{F-Measure evolution for the two kind of datasets, applying  different clustering methods and cut value. The abscissa (X) defines the cut value  and the ordinate (Y) - the F-Measure.
        }
        \label{fig:clustering}
    \vspace*{-3em}
\end{figure}
\vspace*{-0.35cm}
\subsubsection{Tests on different agglomerative clustering methods:} We performed several tests on different AC, namely: single, centroid, average, weighted, complete, ward, and median, that basically vary in the way the distance between cluster elements is estimated \cite{DBLP:journals/corr/abs-1105-0121}. Fig. \ref{fig:clustering} (left) represents the F-Measure of the different clusterings on the SenseCam data, Fig. \ref{fig:clustering} (center) - on the Narrative data and Fig. \ref{fig:clustering} (right) on all data. We can observe that the clustering follows the same behaviour for the two types of data sets, although for the SenseCam the methods are achieving better results than for the Narrative sets. That is reasonable due to the significant difference in image appearance. Despite for the SenseCam sets the complete is achieving the same results as the average methods, the third figure shows how for the whole data the average method is achieving the best results (FM=0.56), followed by the complete (FM=0.55) and the single (FM=0.54). The cut value seems to be very influential for the results since there is a point from which, for each method, clusters all data in one single cluster, leading to FM=0.
\vspace*{-1em}
\begin{figure}
    \vspace{-1em}
        \centering
        \begin{minipage}[b]{0.33\textwidth}
                \includegraphics[width=\textwidth]{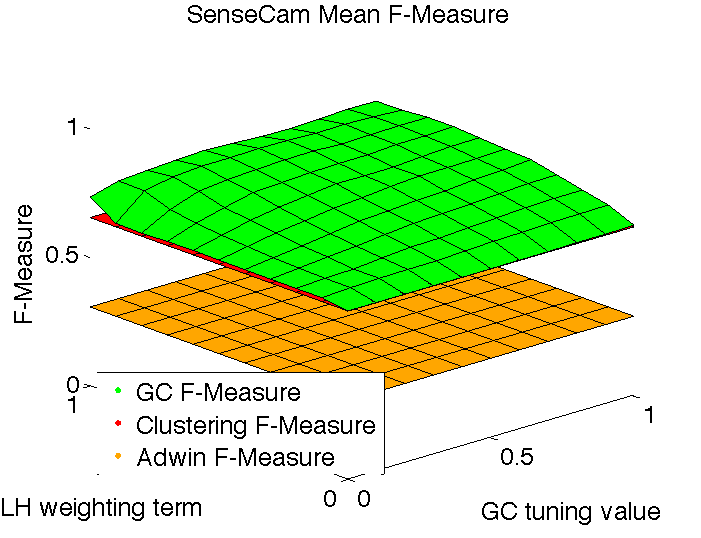}
        \end{minipage}%
        \hspace{-0.5em}
        \begin{minipage}[b]{0.33\textwidth}
                \includegraphics[width=\textwidth]{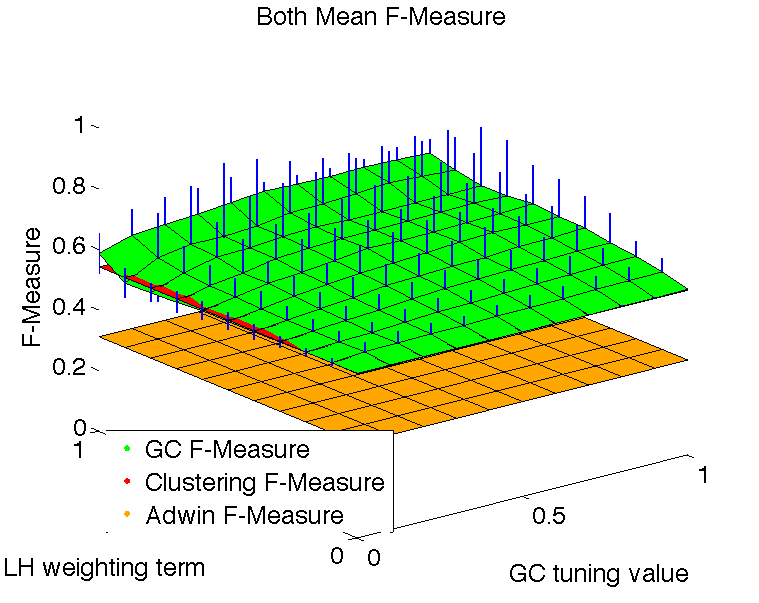}
        \end{minipage}
       \hspace{-0.5em}
       \begin{minipage}[b]{0.33\textwidth}
                \includegraphics[width=\textwidth]{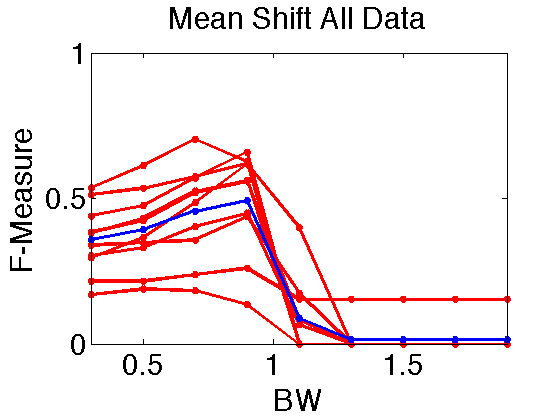}
        \end{minipage}        \vspace{-1em}
        \caption{Average F-Measure of R-Clustering with the best parameters for the SenseCam data (left), and on all datasets (center). The abscissa (X) defines the pair-wise term, the ordinate (Y) the ADWIN vs. AC trade-off and the applicate (Z) shows the corresponding FM. The red surface represents the F-measure of AC and the orange one of ADWIN. MeanShift  performance for video segmentation (right). The abscissa (X) defines the bandwith. The blue line represents the average FM, whereas the red lines are the FM per each dataset.}
        \label{fig:GCclustering}
\end{figure}

\vspace*{-3.5em}
\subsubsection{Tests on R-Clustering using Graph-Cuts:}
\begin{figure}[ht!]
    \vspace{-1em}
    \centering
    \includegraphics[width=1.0\linewidth]{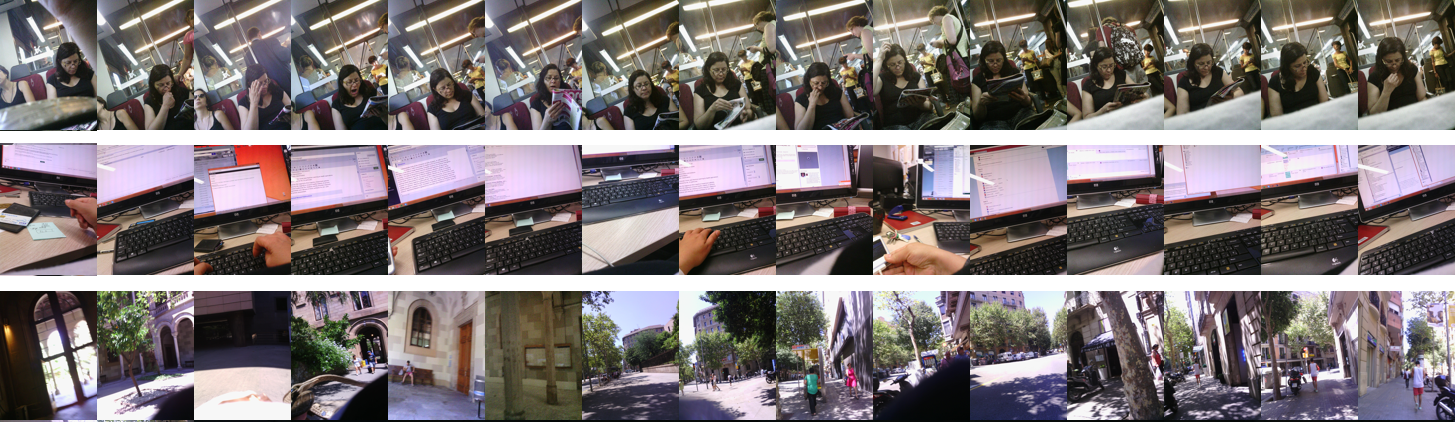}
    \caption{Illustration of our R-Clustering segmentation results for 3 events from a Narrative set.}
    \label{fig:segmentation_results2}
 \vspace{-2em}
\end{figure}

We tested the  R-Clustering performances according to the parameters $\omega_1$ and $\omega_2$. Fig. \ref{fig:GCclustering} (left) shows the average measure on the Sensecam data as a function of both parameters. The optimal F-Measure on all datasets is achieved when $\omega_1=1$ and $\omega_2=0.5$ (Fig. \ref{fig:GCclustering} (center)). Despite the average AC achieves the best performance on our data sets (FM=0.56), the R-Clustering based on this AC method just achieves a FM=0.63. Whereas when it is based on the single clustering, the one that was achieving in AC the second best results (FM=0.54), it achieves the highest FM=0.66 with R-Clustering. Table \ref{tab:results_summary} shows the optimal F-measure for AC, ADWIN and R-Clustering, where the application of R-Clustering method clearly outperforms the F-Measure obtained by the AC and ADWIN technique. Thus, by having $\omega_1=1$ as unary energy parameter proves that the combination of ADWIN (by using its resulting likelihoods and labels initialization) and AC (by using its clusters split in the GC labels initialization) helps to obtain better results by the R-Clustering. In Fig. \ref{fig:GCclustering} (center), the lines depict the standard deviation on each combination of parameters, hence the standard deviation of the best peak results is very low (std=0.17, short line) compared to the higher deviation (longer lines) in the center, meaning that our method is robust and stable. Final video segments can be seen in Fig. \ref{fig:segmentation_results2} that shows three segments corresponding to metro, office and street environments, extracted from a Narrative set.

\vspace*{-0.35cm}
\subsubsection{Tests on other clustering methods:} We compare R-Clustering to K-Means \cite{kmeansClust} and MeanShift (MS) \cite{Fukunaga:2006:EGD:2263309.2268796} (see Fig. \ref{fig:GCclustering} (right) and Table \ref{tab:results_summary}) that achieved FM=0.52 and FM=0.49, respectively. The worse performance can be explained by several facts. The k-Means algorithm requires the number of clusters to be specified and it is not a robust method due to its local minima problem. Considering MeanShift (MS), it is based on density estimation which can deal with arbitrarily shaped data distributions, but its problem is that it is very sensitive to the bandwith (BW) parameter: a large BW can slow down the convergence and a small BW can make it quickly converge leading to over-segmentation. 

\begin{table}[ht!]
\vspace*{-1em}
\centering
\begin{tabular}{c||c|c|c|c|c}
 Datasets & K-Means & Mean-Shift & ADWIN & AC & R-Clustering \\ \hline
 Narrative & 0.32 & 0.38 & 0.32 & 0.45 & \textbf{0.55}  \\
 SenseCam & 0.65 & 0.60 & 0.31 & 0.68 & \textbf{0.79}  \\
 All & 0.52 & 0.49 & 0.31 & 0.56 & \textbf{0.66}  
\end{tabular}
\caption{Average F-Measure result for each of the tested methods on our egocentric datasets.}
\label{tab:results_summary}
\vspace{-1em}
\end{table}

\subsubsection{Application to human tracking for social events characterization:}
\begin{figure}[ht!]
\vspace{-2em}
\includegraphics[width=1.0\linewidth]{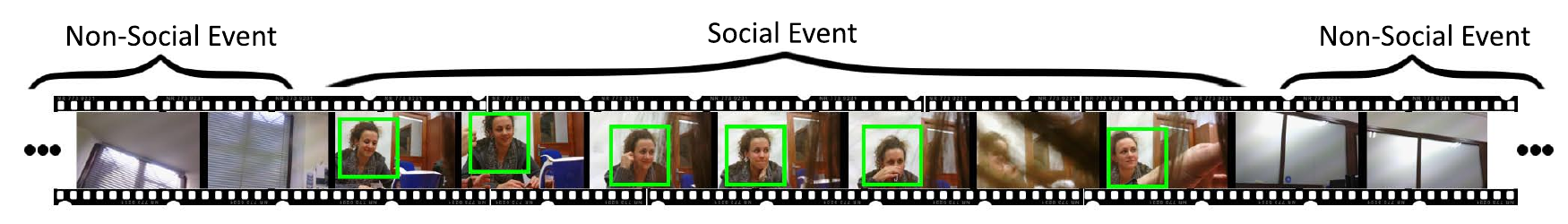}
\caption{Illustration of detecting social events in temporally segmented videos.}
\label{fig:segment}
\vspace{-4em}
\end{figure}

Temporal segmentation is very useful to detect social events, which are characterized  by the presence of people with whom the camera's wearer communicates. Since the presence of people in a specific event likely last from the beginning of the event to its end, social events can be extracted by relying on temporal segmentation. As outlined in \cite{BOT}, due to the substantial difference in frame rate between videos captured by a SenseCam and classical videos, state-of-the-art tracking methods are not directly applicable to lifelogging videos. In \cite{BOT}, the authors  introduced a novel approach, called bag-of-tracklets, that allows to extract robust tracklet prototypes from video segments containing trackable people. While in this work temporal segments are defined manually, we use our detected segments as a pre-processing step for extracting tracklets of people in egocentric videos (Fig. \ref{fig:segment}).
\section{Conclusions}
\label{sec:conclusions}
\vspace*{-0.3cm}In this work, we proposed a novel methodology for automatic egocentric video segmentation that is able to segment low temporal resolution data by global low-level processing. R-Clustering is a robust segmentation approach based on a GC extension technique, that integrates a statistical bound by the concept drift method ADWIN and AC, two methods with complementary properties for temporal video segmentation. We evaluated the performance of R-Clustering on different clustering techniques and on 10 datasets acquired through different wearable devices, and we showed the improvement of the proposed method with respect to the state-of-the-art. 

\section*{Acknowledgments} 
\vspace*{-0.25cm} This work was partially founded by TIN2012-38187-C03-01 and SGR 1219.

\bibliographystyle{plain}
{\footnotesize
\bibliography{Bibliography}}

%
%





\end{document}